%% file: final.tex
\documentclass{article}

\usepackage[final]{corl_2026} 

\usepackage[utf8]{inputenc} 
\usepackage[T1]{fontenc}    
\usepackage{hyperref}       
\usepackage{url}            
\usepackage{booktabs}       
\usepackage{amsfonts}       
\usepackage{nicefrac}       
\usepackage{microtype}      
\usepackage[dvipsnames]{xcolor}         

\usepackage{multicol}
\usepackage{algorithmic}
\usepackage{algorithm}
\usepackage{amsmath}
\usepackage{amssymb}
\usepackage{amsthm}
\usepackage{graphics}
\usepackage{epsfig}
\usepackage{multirow}
\usepackage{wrapfig}
\usepackage{mathtools}
\usepackage{makecell}
\usepackage{bbm}
\usepackage{booktabs}
\usepackage{graphicx}
\usepackage{kantlipsum}
\usepackage{centernot}
\usepackage{xfrac}
\usepackage{afterpage}
\usepackage[dvipsnames,table]{xcolor}
\usepackage{tablefootnote}
\usepackage{enumitem}
\usepackage{bbding}
\usepackage{subcaption}

\definecolor{gred}{rgb}{0.859,0.267,0.216}
\definecolor{ggreen}{rgb}{0.059,0.616,0.345}

\definecolor{stanfordred}{RGB}{140,21,21} 
\hypersetup{
    colorlinks=true,
    linkcolor=stanfordred,
    citecolor=stanfordred,
    urlcolor=stanfordred,
    filecolor=stanfordred,
    pdftitle={Mixture of Frames Policy: Multi-Frame Action Denoising for Bimanual Mobile Manipulation},
    pdfauthor={Dian Wang, Jisang Park, Xiaomeng Xu, Han Zhang, Shuran Song, and Jeannette Bohg},
    pdfsubject={Preprint on robot learning and bimanual mobile manipulation},
    pdfkeywords={bimanual mobile manipulation, diffusion policy, coordinate frames, mixture of experts}
}

\newcommand{\SE}{\mathrm{SE}}
\newcommand{\SO}{\mathrm{SO}}

\usepackage{xparse}
\ExplSyntaxOn
\NewDocumentCommand{\definealphabet}{mmmm}
 {
  \int_step_inline:nnn { `#3 } { `#4 }
   {
    \cs_new_protected:cpx { #1 \char_generate:nn { ##1 }{ 11 } }
     {
      \exp_not:N #2 { \char_generate:nn { ##1 } { 11 } }
     }
   }
 }
\ExplSyntaxOff
\definealphabet{bb}{\mathbb}{A}{Z}
\definealphabet{bf}{\mathbf}{A}{Z}
\definealphabet{mc}{\mathcal}{A}{Z}
\definealphabet{mf}{\mathfrak}{A}{Z}
\definealphabet{mf}{\mathfrak}{a}{z}

\newcommand\blfootnote[1]{%
  \begingroup
  \renewcommand\thefootnote{}\footnote{#1}%
  \addtocounter{footnote}{-1}%
  \endgroup
}

\title{Mixture of Frames Policy: Multi-Frame Action Denoising for Bimanual Mobile Manipulation}

\author{
  Dian Wang$^*$ \enspace Jisang Park$^*$ \enspace  Xiaomeng Xu \enspace Han Zhang \enspace Shuran Song$^\dagger$ \enspace Jeannette Bohg$^\dagger$ \\
  Stanford University\\
  \href{https://mofpo.github.io}{\texttt{https://mofpo.github.io}}
}

\begin{document}
\maketitle


\blfootnote{$^*$ Indicates equal contribution, $^\dagger$ indicates equal advising.}

\vspace{-1cm}
\begin{figure}[h]
\centering
\includegraphics[width=1\linewidth]{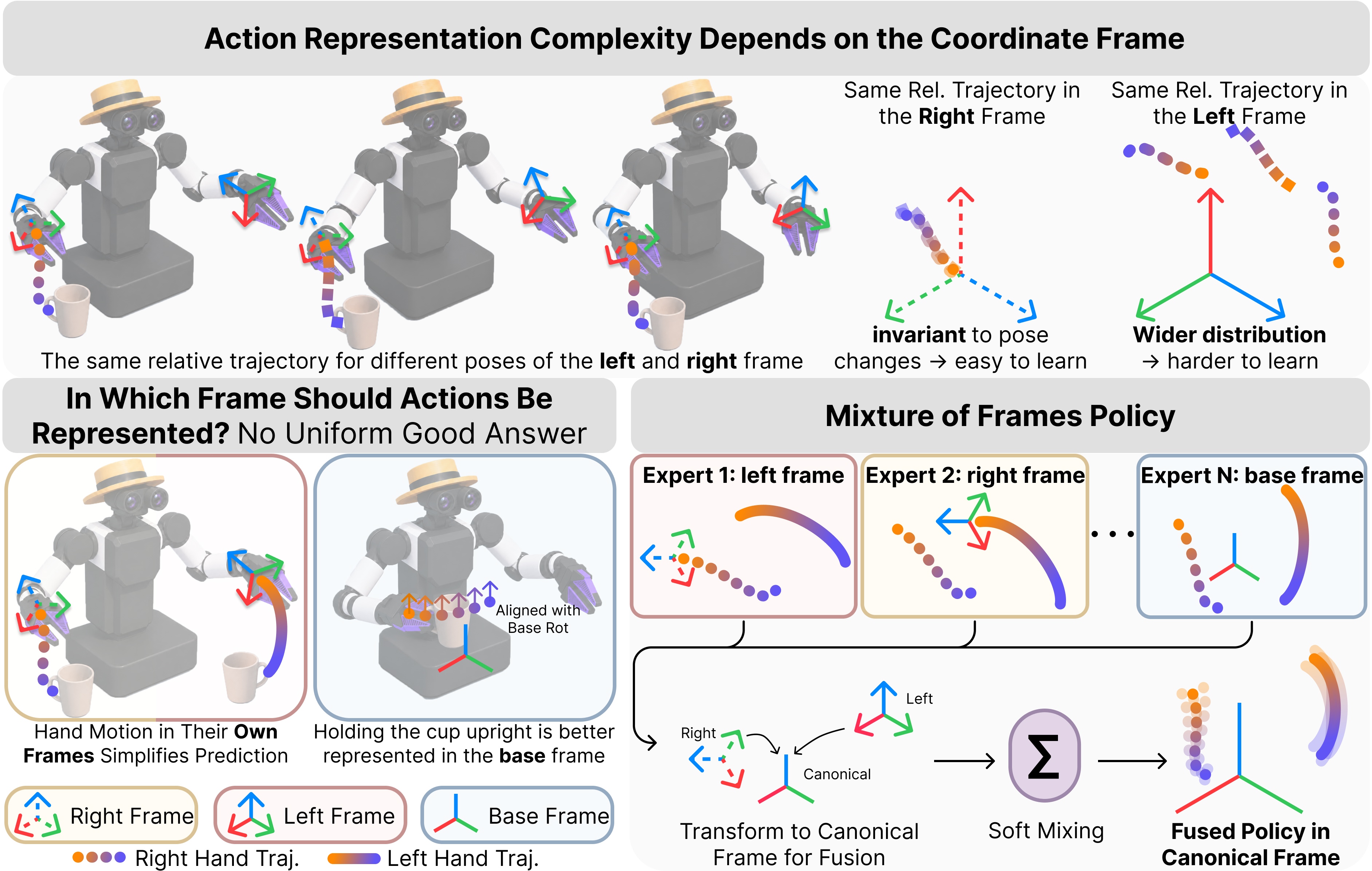}
\caption{\textbf{Mixture of Frames Policy (MoF). } The complexity of a manipulation action distribution depends strongly on the coordinate frame in which it is represented. The same relative motion can be compact and invariant in one frame (e.g. an end-effector frame), yet form a broader and harder-to-learn distribution in another. Conversely, some motions,  such as keeping an object upright, are better represented in a base-aligned frame. Since no single frame is universally optimal, MoF trains frame-specialized experts and fuses their predictions in a shared canonical frame.}
\label{fig:teaser}
\end{figure}

\begin{abstract}
Robotic manipulation is inherently multi-frame: local actions may be simple in an end-effector frame, while transport, upright-object handling, and whole-body coordination are better represented in a base-aligned frame. However, modern diffusion-based visuomotor policies typically commit to a single predefined action frame, forcing one denoiser to model action distributions that are often unnecessarily complex in that frame. We propose Mixture of Frames Policy (MoF), a diffusion policy that performs synchronized action denoising across multiple coordinate frames. MoF maintains a single canonical diffusion state, re-expresses it in several task-relevant frames, applies frame-specialized denoisers, and fuses their noise predictions back in the canonical frame. To make this possible for intermediate noisy diffusion states, we adopt a column-based 6D rotation representation within an SE(3) action parameterization that supports exact, differentiable frame transformations without requiring noisy rotations to lie on the SO(3) manifold. Across nine simulated bimanual manipulation tasks, we show that the best action frame is task-dependent and that MoF improves over  oracle frame selection and standard Mixture-of-Experts (MoE) baselines. We further evaluate MoF on two real-world bimanual mobile manipulation tasks, demonstrating that it outperforms all constituent single-frame baselines. 
\end{abstract}



\section{Introduction}
\label{sec:intro}
Robotic manipulation systems are inherently multi-frame. A bimanual mobile robot naturally operates with a base frame, left and right end-effector frames, camera frames, and other task-relevant coordinate frames defined by its kinematics and embodiment. These frames are not only alternative parameterizations of the same state, but can also make different parts of the same task easier to model. For example, the action of approaching and grasping a cup handle is more naturally expressed in the gripper frame: if the cup and gripper appear in the same relative configuration, the desired gripper frame action is invariant, even if the gripper-object pair is in different absolute poses relative to the base frame (Fig.~\ref{fig:teaser} top). In contrast, when the robot transports a filled cup, maintaining the cup upright and moving it coherently with the mobile base is more naturally expressed in the base frame.  

This frame choice matters because different tasks can be easier to learn when the action is represented in a specific frame. However, modern visuomotor policies such as Diffusion Policies~\cite{chi2025diffusion} typically choose a single predefined coordinate frame to parameterize the action. Even when transformations among frames are provided as proprioceptive observations, the denoiser must still model the entire action distribution in one chosen frame. Moreover, the most useful frame might vary while a manipulation task is progressing. A single-frame policy therefore inherits whatever complexity the chosen frame imposes on the action distribution, regardless of whether that complexity is intrinsic to the task or merely an artifact of the parameterization.

Reasoning over multiple coordinate frames offers a natural solution to this limitation, allowing the policy to use each frame when the corresponding action distribution is easiest to model.
Prior work has explored multi-frame reasoning in robot learning by mining observation frames for point-cloud policies~\cite{liu2023frame}, and composing low-dimensional frame-conditioned skills from demonstrations in non-visual, state-based skill learning~\cite{montero2024learning}. However, multi-frame reasoning has remained largely under-explored in the context of modern diffusion-based visuomotor policies. 

This suggests that frames should not only be provided as observations, but should also define alternative action spaces in which denoising can be performed.
In this paper, \emph{we propose Mixture of Frames policy (MoF), a diffusion policy that denoises actions in multiple coordinate frames in parallel}. Instead of using a single denoiser in one predefined coordinate frame, MoF instantiates one expert denoiser per reference frame and fuses their predictions at each denoising step through either a mixture-of-experts pipeline or simple ensembling. This allows the policy to reason over multiple coordinate frames simultaneously, rather than committing to a fixed frame selected by the designer. 
To our knowledge, MoF is the first diffusion-based visuomotor policy that performs denoising directly across multiple action frames.
Our contributions can be summarized as follows:
\begin{itemize}[leftmargin=6mm]
\item We present a systematic analysis showing that the choice of coordinate frame has a substantial and task-dependent effect on diffusion policy performance in bimanual mobile manipulation, motivating policies that reason natively across multiple frames.
\item We propose Mixture of Frames (MoF) Policy, a diffusion policy architecture that performs synchronized action denoising in multiple coordinate frames in parallel, with per-frame expert denoisers and a learned router or uniform weighting that combines their predictions.
\item We adopt a column-vector $\SE(3)$ action representation that admits exact, differentiable transformation of both noisy and noise-free actions across reference frames, 
facilitating consistent multi-frame denoising.
\end{itemize}
	

\section{Related Works}
\label{sec:related_works}

\paragraph{Learning Bimanual Mobile Manipulation}
Bimanual mobile manipulation requires learning coordinated whole-body motion.
Prior works have addressed bimanual coordination by decomposing roles~\cite{grannen2023stabilize} or modeling inter-arm interactions~\cite{lee2025interact, jiang2025rethinking, chen2026rotri}, while mobile manipulation approaches often decouple whole-body coordination from policy learning~\cite{yang2025mobi, ha2025umi, sundaresan2025homer}. Most recent bimanual mobile manipulation works address both via co-training with fixed-base data~\cite{fu2025mobile, zhu2026emma} or introducing specialized architectures~\cite{jiang2025behavior, xu2026hommi}. However, these approaches typically rely on a single frame representation, which may be insufficient when coordination requirements shift throughout task progression.

\paragraph{Multi-Frame Reasoning for Manipulation}
Manipulation policies benefit from multi-frame reasoning as different decisions are most naturally expressed in distinct coordinate systems~\cite{2014_ARMProject_USC}. Earlier task-parameterized methods address this by composing multiple local frames via generalized weighting~\cite{huang2018generalized} or time-varying relevance estimation~\cite{montero2024learning}. Subsequent learning-based works introduce adaptive frame selection in point-cloud policies~\cite{liu2023frame} and frame-based interfaces in hierarchical control~\cite{zhao2025hierarchical,zhao2026generalizable}. In parallel, recent bimanual manipulation works adopt relation-centric representations to encode hand-object and arm-arm geometries~\cite{gao2024bi, bahety2024screwmimic, chen2026rotri}. 
These studies motivate the adaptive fusion of frame-specialized experts, as optimal frames vary by subproblem.

\paragraph{Mixture-of-Experts for Manipulation}
Mixture-of-Experts (MoE)~\cite{jacobs1991adaptive, shazeer2017outrageously} enables modular experts to capture diverse local behaviors more effectively than monolithic policies in robot learning~\cite{celik2022specializing, celik2024acquiring}. Existing MoE-based manipulation policies specialize experts along behavioral dimensions such as task identity~\cite{wang2025sparse, guo2026moe}, skill abstraction~\cite{hao2026abstracting, rodriguez2026lar}, and execution phases~\cite{cheng2026moe}, or architectural components such as denoising stages~\cite{reussefficient} and sensing modalities~\cite{yuforcevla}. We instead formulate MoE over coordinate frames, addressing the underexplored challenge of multi-frame reasoning through the adaptive fusion of frame-specific action representations.


\section{Method}
\label{sec:method}

We present \textbf{Mixture of Frames} (MoF) Policy, a diffusion policy that denoises a single action trajectory through multiple coordinate-frame parameterizations.
In this section, we first present the problem statement, then introduce a synchronized multi-frame denoising procedure, followed by a transformation-compatible action representation that makes this procedure exact for noisy actions.

\subsection{Problem Statement}
\label{subsec:dp_frames}

We consider visuomotor imitation learning for bimanual mobile manipulation with diffusion policies~\cite{chi2025diffusion}. At each control step, the policy receives an observation $o=(\mathcal{I},q)$, where $\mathcal{I}$ denotes RGB images and $q$ denotes proprioception. The policy predicts an action chunk $a\in\mathbb{R}^{\tau\times d}$ over horizon $\tau$. In our setting, each action timestep contains the target poses of the left and right end-effectors and the gripper commands. These end-effector targets are tracked by a whole-body controller for execution. We denote $\mathcal{F}=\{F_m\}$ as a family of task-relevant coordinate frames, including but not limited to the left and right end-effector frames $F_l$ and $F_r$, and the base frame $F_b$. Given proprioception, the rigid transforms from any frame $m$ to another frame $m'$, $^{m'}T_{m} \in \mathrm{SE}(3)$, are known.

In standard diffusion policies, the designer fixes a reference action frame $F^\star\in\mathcal{F}$, and both training targets and denoising predictions are represented in this frame. For example, prior works have used the world frame~\cite{chi2025diffusion,wang2024equivariant,XuX-RSS-25}, the end-effector frame~\cite{wang2025practical,xu2026compliant}, or the left-hand frame~\cite{xu2026hommi}. However, it is unclear which single fixed $F^\star$ is universally best for bimanual mobile manipulation. 
This motivates a policy that maintains multiple frame-specialized denoisers and combines their predictions adaptively.

\subsection{Consistent Multi-Frame Denoising}
\label{sec:multi_frame_denoising}

\begin{figure}[t]
\centering
\includegraphics[width=\linewidth]{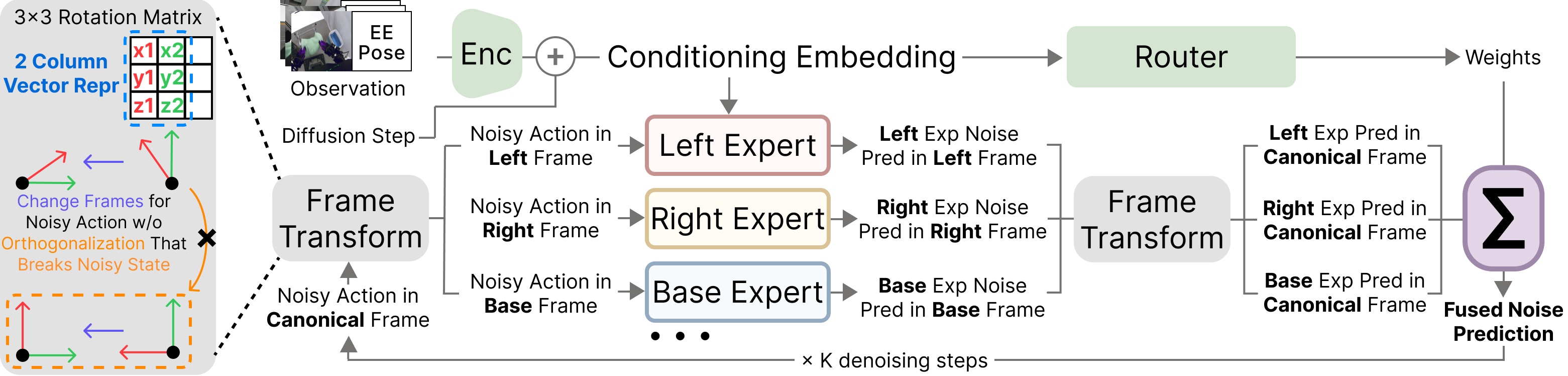}
\caption{
\textbf{MoF Architecture.} At each diffusion step, MoF maintains a noisy action in the canonical frame. This noisy action is re-expressed in each expert's frame, and all experts predict noise in their native frame. 
The predicted noise vectors are then transformed back to the canonical frame and fused using either router-predicted weights or fixed uniform weights.
The fused noise prediction is used by a standard denoising step, keeping all frame experts synchronized along the same diffusion trajectory. Left: to enable frame change for noisy actions, we employ a column-vector action representation where frame transform can be achieved without orthogonalization, preserving the noisy action state.
}
\label{fig:architecture}
\end{figure}

A straightforward way to use multiple frame-specialized denoisers (i.e., experts) would be to run an independent diffusion process in each frame and fuse the final denoised actions. However, this is problematic with multimodal action distributions, as different denoisers may converge to different modes, making their final action predictions incompatible. Instead, MoF keeps all experts synchronized by maintaining a single diffusion trajectory in a canonical frame and fusing predictions at every denoising step, as shown in Fig.~\ref{fig:architecture}.

Specifically, let $F_c$ denote the canonical frame in which the scheduler state $x^k_c$ is stored. For each frame $F_m\in\mathcal{F}$, 
we transform the canonical-frame noisy action sample \(x^k_c\) into the expert frame via ${}^{m}\mathcal{T}_{c}(\cdot)$, feed it to a frame-specialized denoiser $\epsilon^{\theta}_m$, transform the predicted noise back to the canonical frame using ${}^{c}\mathbf{T}_{m}(\cdot)$, and fuse the aligned predictions:
\begin{equation}
\hat{\epsilon}_c
=
\sum_{F_m\in\mathcal{F}}
w_m\,
{}^{c}\mathbf{T}_{m}
\!\left(
\epsilon^{\theta}_m
\left(
{}^{m}\mathcal{T}_{c}(x^k_c), o; k
\right)
\right).
\label{eq:mof_fused_noise}
\end{equation}
Here, both $\mathcal{T}$ and $\mathbf{T}$ are induced by the rigid transform between $F_c$ and $F_m$, and their concrete definitions are given in Sec.~\ref{subsec:action_rep}. $w_m$ is the weight for the denoiser in frame $m$, which can be either set by the designer or learned through a router, with $\sum_m w_m=1$. 
This synchronized update has two benefits. First, all experts contribute to the same underlying diffusion trajectory, avoiding the mode-mismatch problem. Second, because fusion occurs in the canonical noise space, the diffusion scheduler remains unchanged: MoF can be implemented on top of a standard DDIM sampler by replacing the single noise prediction with the fused multi-frame prediction $\hat{\epsilon}_c$.

Training follows the same principle. We sample a canonical-frame action $x^0_c=a$, noise $\epsilon_c\sim\mathcal{N}(0,I)$, and diffusion step $k$, construct $x^k_c$ the same way as standard Diffusion Policy, and train the fused canonical prediction to match the canonical noise:
\begin{equation}
\mathcal{L}_{\mathrm{mix}}(\theta)
=
\mathbb{E}_{k,\epsilon_c,o}
\left[
\left\|
\epsilon_c
-
\sum_{F_m\in\mathcal{F}}
w_m\,
{}^{c}\mathbf{T}_{m}
\!\left(
\epsilon^{\theta}_m
\left(
{}^{m}\mathcal{T}_{c}(x^k_c),o;k
\right)
\right)
\right\|^2
\right].
\label{eq:mof_loss}
\end{equation}
Thus, the training target remains the standard diffusion noise in the canonical frame, while each expert learns to denoise the same underlying noisy action expressed in its own frame.

\subsection{Transformation-Compatible Action Representation}
\label{subsec:action_rep}

The synchronized denoising procedure in Sec.~\ref{sec:multi_frame_denoising} requires frame transformations to be applied directly to intermediate diffusion states, whose rotation channels generally do not form a valid element of $\SO(3)$. This poses a challenge for the row-vector 6D rotation representation used in standard Diffusion Policy implementations, which stores the first two rows of the rotation matrix. Changing frames under this representation requires reconstructing a valid rotation matrix via orthogonalization before applying the transform. This projection is nonlinear and lossy: it replaces the noisy rotation channels with a projected rotation, thereby changing the diffusion state itself.

We instead use a column-vector rotation representation. For one robot arm at one action timestep, the pose/action channels expressed in the canonical frame \(F_c\) are \(x_c=[p_c,c^1_c,c^2_c,g]\), where \(p_c\in\mathbb{R}^3\) is the end-effector position, \(c^1_c,c^2_c\in\mathbb{R}^3\) are the first two columns of the end-effector orientation, and \(g\) is the gripper command. For a clean end-effector orientation, let \({}^{c}R_{ee}=[c^1_c,c^2_c,c^3_c]\) denote the rotation from the end-effector frame to the canonical frame. Re-expressing the same orientation in an expert frame \(F_m\) uses the known frame-change rotation \({}^{m}R_c\), giving \({}^{m}R_{ee}={}^{m}R_c{}^{c}R_{ee}=[{}^{m}R_cc^1_c,{}^{m}R_cc^2_c,{}^{m}R_cc^3_c]\). Thus, each stored column transforms as \(c^i_m={}^{m}R_cc^i_c\). This operation is well-defined even when \(c^1_c\) and \(c^2_c\) are arbitrary noisy vectors rather than columns of a valid rotation matrix. Therefore, changing action frames only requires left-multiplying ordinary 3D vectors by the frame-change rotation \({}^{m}R_c\), without projecting onto \(\SO(3)\), as illustrated in Fig.~\ref{fig:architecture} left.

Let \({}^{m}T_c\) denote the rigid transform from the canonical frame \(F_c\) to an expert frame \(F_m\). This transform induces the action-state operator \({}^{m}\mathcal{T}_c\) and the noise-vector operator \({}^{m}\mathbf{T}_c\) used in Eq.~\eqref{eq:mof_fused_noise}. Their explicit definitions differ only in whether the translation offset is applied to position channels and are provided in Appendix~\ref{app:transforms}. With this representation, MoF can transform noisy action states and predicted noise vectors across frames without projecting noisy rotations back onto $\SO(3)$.


\section{Simulation Experiments}
\label{sec:sim_exp}
\vspace{-.5\baselineskip}

\begin{figure}[t]
    \centering
    \includegraphics[width=\linewidth]{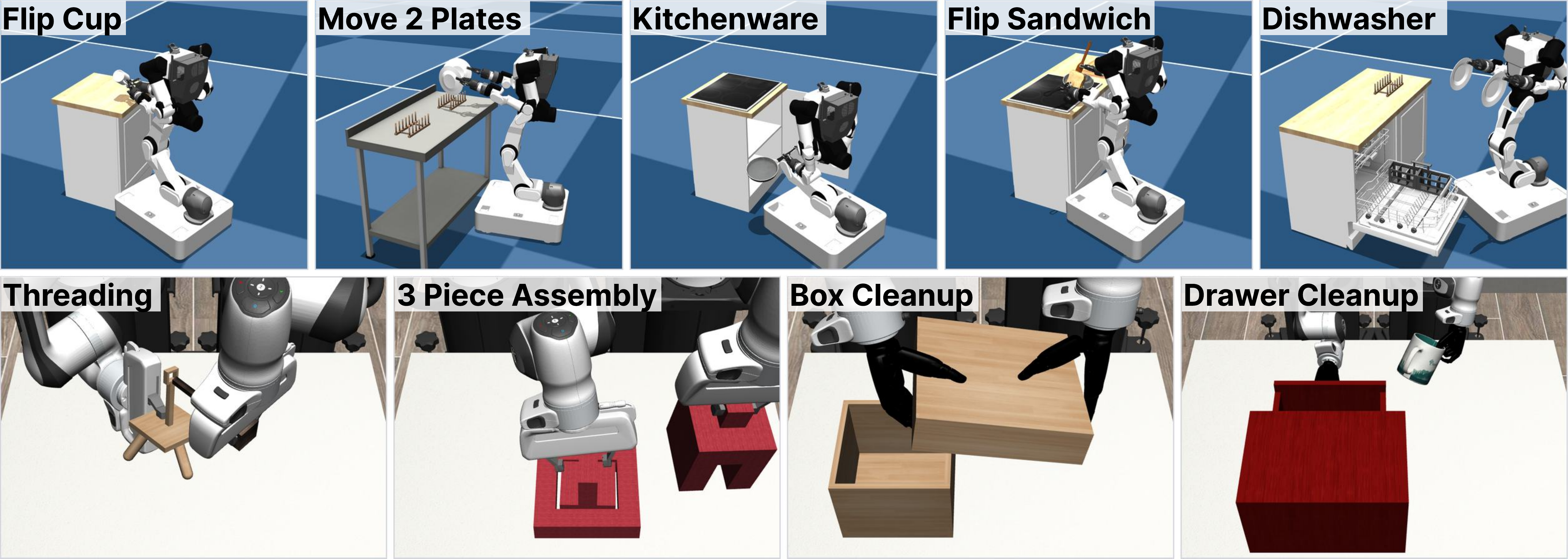} 
    \caption{\textbf{Simulation Tasks.} Nine tasks from BiGym~\cite{chernyadev2025bigym} (top) and DexMimicGen~\cite{jiang2025dexmimicgen} (bottom).}
    \label{fig:sim_experiment}
\end{figure}

\begin{table}[t]
\centering
\caption{Single-frame DP baselines (\%, mean $\pm$ std.\ err.\ over 3 seeds).  \textbf{Oracle} selects the best frame per task (bold). \textbf{Worst} selects the worst per task, with parenthetical values denoting Worst $-$ Oracle.}
\label{tab:single_frame_baselines}
\tiny
\setlength{\tabcolsep}{0.5pt}
\begin{tabular}{l c c c c c c c c c c}
\toprule
Frame & Avg & Flip Cup & Move 2 Plates & Kitchenware & Flip Sandwich & Dishwasher & Threading & 3 Piece Assembly & Box Cleanup & Drawer Cleanup \\
\midrule
Left & 55.6 & $47.3\pm1.0$ & $44.5\pm3.1$ & $29.2\pm3.5$ & $\mathbf{36.1\pm3.3}$ & $78.8\pm2.5$ & $43.9\pm4.4$ & $51.5\pm6.8$ & $\mathbf{92.4\pm1.2}$ & $76.8\pm1.7$ \\
Right & 55.0 & $45.1\pm1.9$ & $\mathbf{47.2\pm1.7}$ & $22.5\pm0.7$ & $34.3\pm1.6$ & $83.1\pm0.5$ & $34.7\pm3.5$ & $69.1\pm2.9$ & $79.9\pm7.2$ & $79.6\pm1.5$ \\
Base & 61.0 & $\mathbf{53.9\pm2.9}$ & $29.7\pm0.1$ & $\mathbf{31.2\pm2.6}$ & $35.5\pm0.3$ & $\mathbf{84.1\pm1.6}$ & $62.3\pm3.0$ & $\mathbf{76.3\pm0.5}$ & $86.1\pm6.3$ & $\mathbf{90.1\pm1.0}$ \\
Rel Traj & 55.8 & $42.0\pm2.1$ & $32.1\pm2.6$ & $26.1\pm1.9$ & $25.5\pm3.2$ & $76.4\pm0.6$ & $\mathbf{63.1\pm0.7}$ & $73.7\pm2.3$ & $80.0\pm2.3$ & $83.5\pm1.0$ \\
\midrule
Oracle Frame & 63.8 & 53.9 & 47.2 & 31.2 & 36.1 & 84.1 & 63.1 & 76.3 & 92.4 & 90.1 \\
Worst Frame & 48.8 (-15.0) & 42.0 (-11.9) & 29.7 (-17.5) & 22.5 (-8.7) & 25.5 (-10.6) & 76.4 (-7.7) & 34.7 (-28.4) & 51.5 (-24.8) & 79.9 (-12.5) & 76.8 (-13.3) \\
\bottomrule
\end{tabular}
\end{table}

Our simulation experiments aim to answer the following questions:
\vspace{-.25\baselineskip}
\begin{enumerate}[
    label=\textbf{Q\arabic*.},
    leftmargin=9mm,
    itemsep=0pt,
    parsep=0pt,
    topsep=0pt,
    partopsep=0pt
]
\item How much does the choice of action frame affect diffusion-policy performance?
\item Can MoF match the best single-frame policy and outperform existing baselines?
\item Which components of MoF are responsible for the performance improvement? 
\end{enumerate}
\vspace{-.25\baselineskip}
We evaluate MoF on nine tasks across BiGym~\cite{chernyadev2025bigym} (\textit{FlipCup}, \textit{Move2Plates}, \textit{Kitchenware}, \textit{FlipSandwich}, \textit{Dishwasher}) and DexMimicGen~\cite{jiang2025dexmimicgen} (\textit{Threading}, \textit{3PieceAssembly}, \textit{BoxCleanup}, \textit{DrawerCleanup}).
All policies are trained for 500 epochs with 100 demonstrations and three random seeds. For each seed, we evaluate five checkpoints from epochs 460--500 at 10-epoch intervals using 50 episodes each. We report the mean success rate (\%) and standard error. See Appendix~\ref{app:sim_details} for details.

\vspace{-.5\baselineskip}
\paragraph{Q1. Evaluation of Frame Choice}
We first ask whether action-frame choice is an important design decision for diffusion policies, and whether a single hand-designed frame is sufficient across tasks. To answer this, we train single-frame diffusion policies whose action representation is fixed to one of the four candidate expert parameterizations used by MoF (See Appendix~\ref{subsec:frames} for details): left end-effector, right end-effector, base-relative, and per-arm relative-trajectory. We also report two aggregate references. Oracle Frame selects the best single frame for each task after evaluation and therefore represents the best possible task-level frame selection among our candidates. Worst Frame selects the worst single frame for each task, quantifying the cost of an unfavorable frame choice.

Table~\ref{tab:single_frame_baselines} shows that action frame selection has a large and task-dependent impact on diffusion policy performance. Among the individual frames, the base frame obtains the highest average success rate, but it is not uniformly best across tasks. It performs well on five tasks, while the left frame is best on two tasks, the right frame and the relative trajectory frame are each best on one task. The gap between the oracle and worst frame further quantifies this sensitivity. On average, the worst frame is 15\%p below the oracle frame.
These results confirm that no single hand-designed frame consistently captures the easiest action distribution across all bimanual manipulation tasks.

\vspace{-.5\baselineskip}
\paragraph{Q2. Comparison of MoF and Baselines}

We next ask whether our method improves over single-frame baselines that do not reason over action frames. 
We evaluate two variants of MoF. \textbf{MoF-MoE}, shown in Fig.~\ref{fig:architecture}, uses a router conditioned on the diffuser-conditioning embedding to predict the expert weights \(w_m\). \textbf{MoF-Ensemble} removes the learned router and uses uniform weights, \(w_m=1/|\mathcal{F}|\), effectively averaging the aligned predictions from all frame experts.
We compare MoF-MoE and MoF-Ensemble with six baselines. \textbf{Oracle Frame} is the strongest single-frame reference from Table~\ref{tab:single_frame_baselines}. Comparing against it tests whether MoF automates frame selection or potentially does more than that. \textbf{DP}~\cite{chi2025diffusion} is the standard Diffusion Policy trained in a single world-frame action representation, measuring improvement over the conventional design. \textbf{MoE-DP}~\cite{cheng2026moe} uses a mixture-of-experts module in the conditioning pathway of DP, but still denoises actions in a single action representation. Comparing against it tests whether the gains come from frame-level denoising rather than a generic MoE architecture. 
We compare against \textbf{Single-Frame Ensemble} and \textbf{MF-DP} to assess the need for frame-specialized denoisers: the former averages four same-frame diffusers, while the latter uses a capacity-matched single denoiser with multi-frame inputs.
\begin{wrapfigure}[15]{r}{0.4\linewidth}
\centering
\vspace{-0.35cm}
\includegraphics[width=\linewidth]{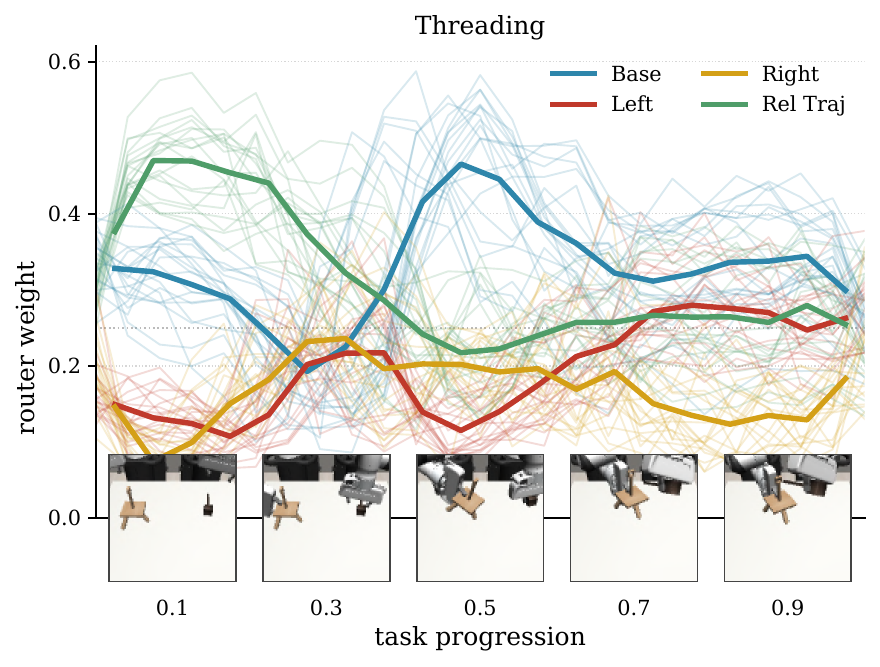}
\vspace{-0.5cm}
\caption{\textbf{Router Weights on Threading.} For each MoF-MoE expert, faint lines show router weights across task progression for individual episodes, and bold lines show the 20-episode mean.
}
\label{fig:threading_router}
\end{wrapfigure}
In the end, we compare against \textbf{Sym-in-DP}~\cite{wang2025practical}, a geometry-aware baseline leveraging rotational equivariance in the vision encoder. 

\begin{table}[t]
\centering
\caption{MoF vs.\ baselines: last-5 success rate (\%, mean $\pm$ std.\ err.\ over 3 seeds). Bold indicates best, underline indicates second.}
\label{tab:moe_vs_baselines}
\tiny
\setlength{\tabcolsep}{0.8pt}
\begin{tabular}{l c c c c c c c c c c}
\toprule
Method & Avg & Flip Cup & Move 2 Plates & Kitchenware & Flip Sandwich & Dishwasher & Threading & 3 Piece Assembly & Box Cleanup & Drawer Cleanup \\
\midrule
MoF MoE (Ours) & \textbf{66.8} & $\underline{56.5\pm1.4}$ & $\mathbf{51.6\pm4.4}$ & $\underline{31.6\pm2.4}$ & $\underline{40.0\pm3.1}$ & $89.2\pm3.9$ & $\underline{70.8\pm7.5}$ & $\underline{77.3\pm2.0}$ & $\underline{87.1\pm3.2}$ & $\mathbf{96.9\pm0.7}$ \\
MoF Ens. (Ours) & \underline{65.9} & $\mathbf{59.2\pm0.4}$ & $\underline{51.5\pm1.5}$ & $\mathbf{33.7\pm0.5}$ & $\mathbf{43.2\pm2.0}$ & $\mathbf{89.7\pm1.4}$ & $68.7\pm3.3$ & $76.0\pm1.3$ & $77.5\pm5.2$ & $93.3\pm2.9$ \\
\midrule
Oracle Frame & 63.8 & $53.9\pm2.9$ & $47.2\pm1.7$ & $31.2\pm2.6$ & $36.1\pm3.3$ & $84.1\pm1.6$ & $63.1\pm0.7$ & $76.3\pm0.5$ & $\mathbf{92.4\pm1.2}$ & $90.1\pm1.0$ \\
MF-DP & 60.3 & $54.0\pm1.6$ & $34.5\pm5.1$ & $22.7\pm3.8$ & $38.8\pm2.4$ & $\underline{89.6\pm2.2}$ & $69.7\pm0.7$ & $70.0\pm2.4$ & $68.4\pm2.1$ & $\underline{95.1\pm0.7}$ \\
Single-Frame Ens. & 55.9 & $46.4 \pm 3.6$ & $31.2 \pm 2.7$ & $25.9 \pm 3.1$ & $31.2 \pm 2.8$ & $74.5 \pm 3.6$ & $69.7 \pm 2.2$ & $75.7 \pm 1.7$ & $66.5 \pm 3.7$ & $81.7 \pm 2.7$ \\
\midrule
Sym-in-DP~\cite{wang2025practical} & 57.6 & $46.3\pm1.9$ & $31.7\pm1.1$ & $26.8\pm1.6$ & $24.3\pm3.7$ & $81.7\pm2.1$ & $63.1\pm4.2$ & $\mathbf{78.4\pm2.3}$ & $79.9\pm4.4$ & $86.5\pm1.5$ \\
MoE-DP~\cite{cheng2026moe} & 55.1 & $32.9\pm3.8$ & $27.9\pm1.7$ & $18.7\pm0.8$ & $29.9\pm2.0$ & $72.7\pm0.9$ & $\mathbf{72.5\pm5.2}$ & $73.5\pm2.6$ & $74.7\pm6.0$ & $93.5\pm1.9$ \\
DP~\cite{chi2025diffusion} & 50.3 & $17.2\pm2.8$ & $38.5\pm0.4$ & $22.0\pm1.6$ & $32.5\pm3.1$ & $61.5\pm2.1$ & $47.9\pm1.9$ & $66.7\pm1.5$ & $73.3\pm8.9$ & $92.9\pm0.9$ \\
\bottomrule
\end{tabular}
\vspace{-2\baselineskip}
\end{table}

Table~\ref{tab:moe_vs_baselines} shows that both MoF variants outperform the single-frame baselines on average. MoF-MoE achieves an average success rate of 66.8\%, improving over MoE-DP by 11.7 percentage points and over standard DP by 16.5 percentage points. MoF-Ensemble achieves a similar average success rate of 65.9\%, while the same-frame Ensemble baseline performs substantially worse. This indicates that multi-frame denoising, rather than simply using multiple diffusers, is the main source of improvement. MoF-MoE also exceeds capacity-matched MF-DP by 6.5 percentage points, showing that multi-frame inputs alone are insufficient and that frame-specialized denoisers are needed to fully exploit multiple action representations. MoF-MoE also outperforms geometry-aware Sym-in-DP by 9.2 percentage points, supporting the benefit of frame-specialized denoising.

MoF also matches or outperforms Oracle Frame on most tasks. The largest improvement is on Threading ($+7.7$ points). To probe where this gain comes from, we run the trained MoF-MoE policy over 20 training demonstrations and record the router's expert weights at each denoising step. 
Fig.~\ref{fig:threading_router} plots these weights, averaged over the last 4 denoising steps, as a function of task progression. The rel-traj expert is the dominant frame during the first part of the episode, where each gripper reaches the two objects, a motion that is naturally compact in an arm-centric frame. The base-relative expert takes over for the second part, during the coordinated bimanual insertion that requires consistent reasoning across both arms. 
This phase-dependent frame usage is not available to any single-frame baseline, and is consistent with MoF-MoE exceeding even Oracle Frame on this task. (See Appendix~\ref{app:router_analysis})

\vspace{-.5\baselineskip}
\paragraph{Q3. Ablation Study}
\begin{wraptable}[12]{r}{0.26\textwidth}
\vspace{-0.4cm}
\centering
\caption{Ablation avg. over four tasks. Parentheses denote differences from ours.}
\vspace{-0.2cm}
\label{tab:mof_ablation_avg}
\tiny
\setlength{\tabcolsep}{3pt}
\begin{tabular}{l c}
\toprule
Method & Avg \\
\midrule
MoF MoE (Ours) & 66.5 \\
\quad Rel Traj as Canonical & 64.0 ($-2.5$) \\
\quad Left as Canonical & 65.8 ($-0.7$) \\
\quad Right as Canonical & 66.4 ($-0.1$) \\
\quad w/o Best Frame & 60.3 ($-6.2$) \\
\quad w/ Ortho Proj & 59.5 ($-7.0$) \\
\midrule
MoF Ensemble (Ours) & 64.2 \\
\quad w/ Ortho Proj & 0.0 ($-64.2$) \\
\bottomrule
\end{tabular}
\vspace{-1.0\baselineskip}
\end{wraptable}
Finally, we ablate MoF design choices on four representative tasks: \textit{FlipCup}, \textit{Move2Plates}, \textit{Threading}, and \textit{BoxCleanup}. Table~\ref{tab:mof_ablation_avg} shows the average success rates. Changing the canonical frame has only a modest effect, with the largest drop being 2.5 points, suggesting that MoF is robust to the arbitrary choice of canonical frame. In contrast, removing the best-performing single frame from the expert set reduces performance by 6.2 points, indicating that the expert frame set is not redundant. Applying orthogonalization before frame transformation reduces MoF-MoE by 7.0 points and collapses MoF-Ensemble to 0.0\%, validating the need for the transformation-compatible action representation in Sec.~\ref{subsec:action_rep}. See Appendix~\ref{app:ablation} for full results and Appendix~\ref{app:frame_scalability} for additional experiments on expert frame selection and scalability.

\begin{figure}
\centering
\includegraphics[width=\linewidth]{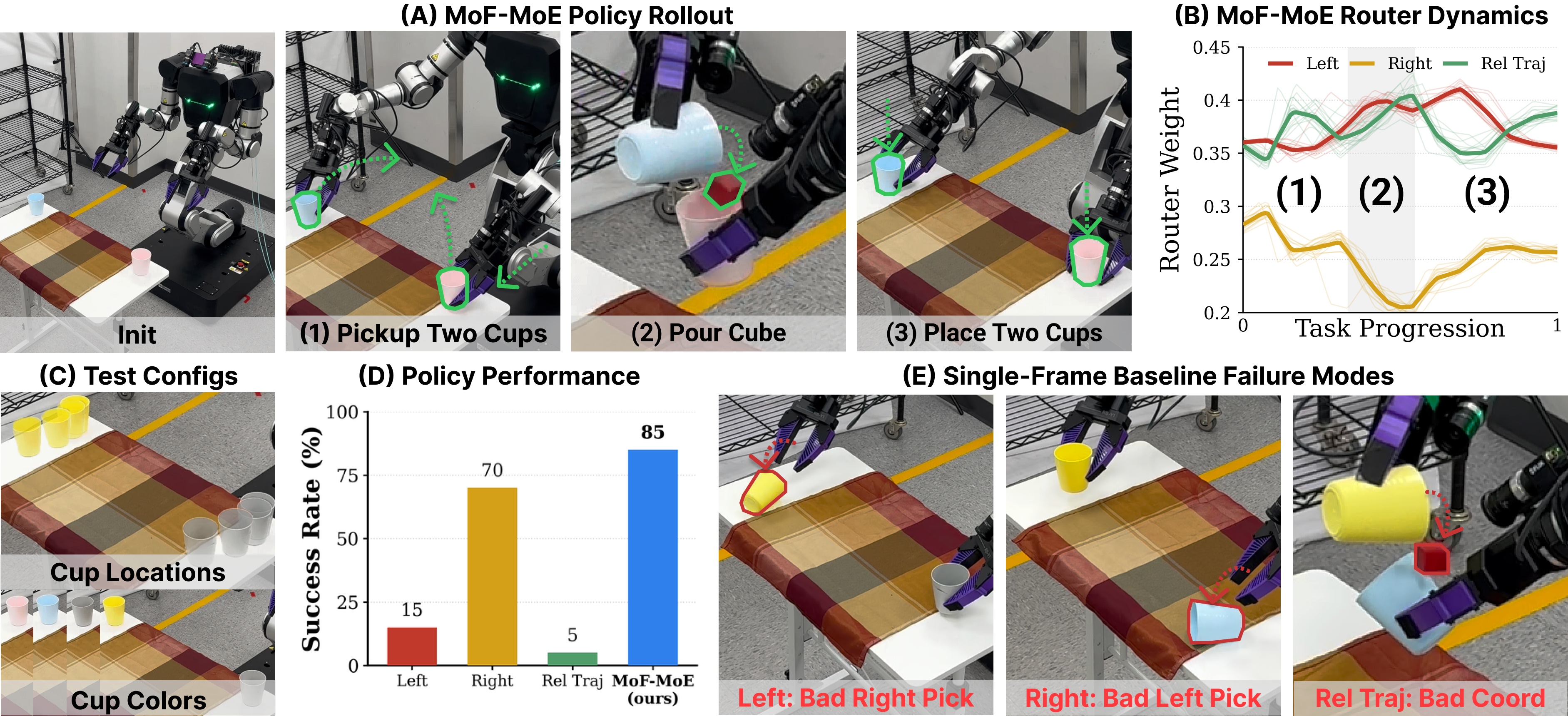}
\caption{\textbf{Pouring Task.} (a) Our MoF policy rollout, (b) average router weights across successful rollouts relative to task progression, (c) evaluation configurations with different cup locations and color combinations, (d) overall performance, and (e) failure cases of each single-frame baseline.}
\label{fig:real_pouring}
\end{figure}

\section{Real World Experiments}
\label{sec:real_exp}

\vspace{-.5\baselineskip}

\paragraph{Setup} Our real-world experiments study how multi-frame fusion improves robustness over single-frame policies in bimanual mobile manipulation. Specifically, our tasks require dynamic switching between multi-frame reasoning modes: global navigation, bimanual coordination, and decoupled manipulation. We integrate MoF-MoE into the HoMMI~\cite{xu2026hommi} real-world framework. As the HoMMI data collection interface lacks base frame tracking, our MoF-MoE policy uses three experts—the left, right, and relative-trajectory experts—and we compare it against the corresponding single-frame baselines. For each task, we evaluate each method over 20 rollouts with different initial configurations and report task success rates. See Appendix~\ref{app:real_details} for details and deeper analysis of results.

\vspace{-.5\baselineskip}

\paragraph{Pouring Task} As shown in Fig.~\ref{fig:real_pouring}, the robot must approach and pick up two cups, pour a cube from the right to the left cup, and place both upright on the table. This requires transitioning between local end-effector-centric reasoning for precise pick-and-place and tight bimanual coordination for pouring. Each policy is trained on 257 demonstrations and evaluated in 20 rollouts across 5 color combinations (3 seen, 2 unseen) and 4 cup placements. Our~\texttt{MoF-MoE} achieves 85\% success, with only 3 failures in picking up (1 left, 2 right). Router analysis reveals \texttt{Left} and \texttt{Rel Traj} alternating during pick-and-place, with \texttt{Left} dominating during pouring as the static left cup provides a more stable reference than the moving right arm. Single-frame baselines fail at distinct phases: (1)~\texttt{Left} (15\%) misses right cup pickup and fails during pouring. (2)~\texttt{Right} (70\%) fails left cup pickup and placement. (3)~\texttt{Rel Traj} (5\%) struggles with left cup pickup or pouring due to unstable grasp poses.

\begin{figure}
\centering
\includegraphics[width=\linewidth]{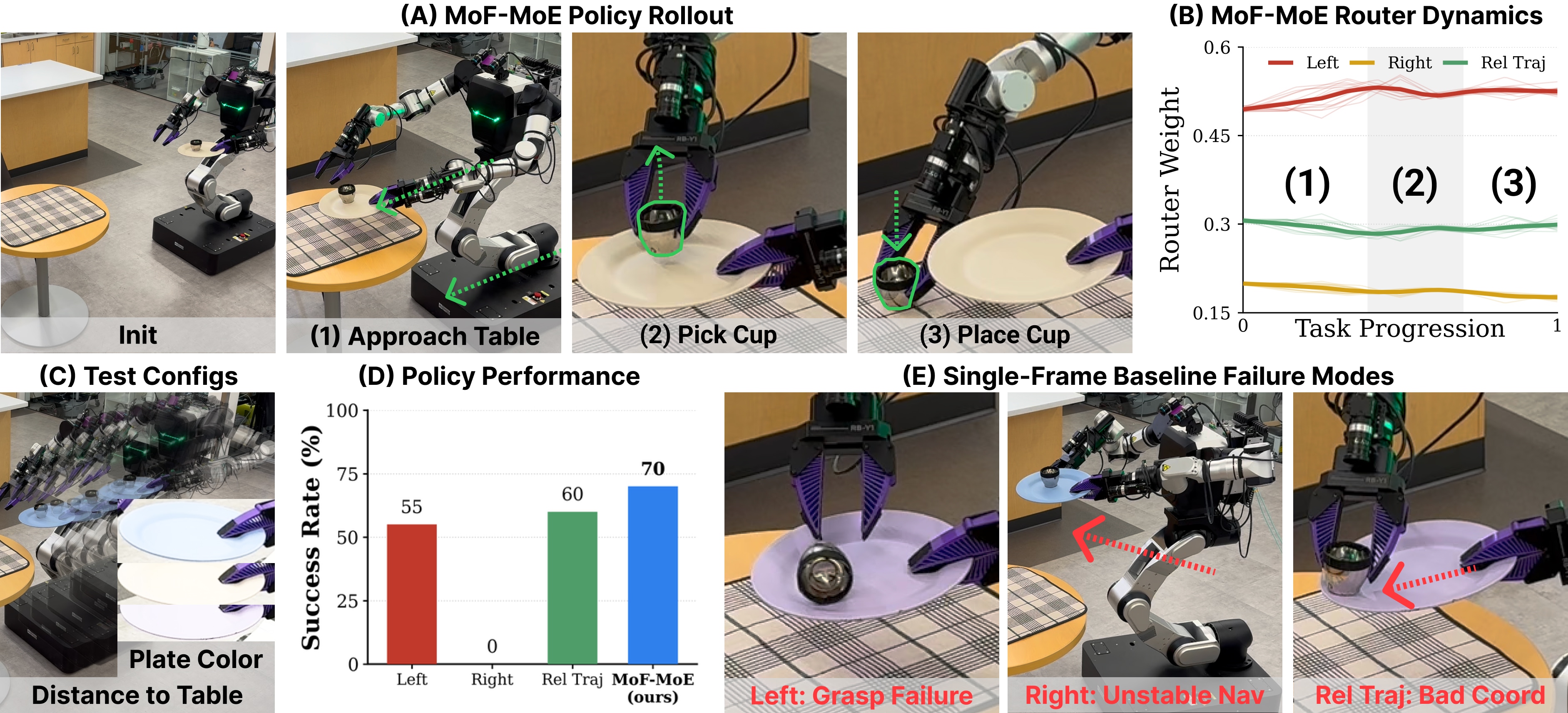}
\caption{\textbf{Serving Task.} (a) Our MoF policy rollout, (b) average router weights across successful rollouts relative to task progression, (c) evaluation configurations with varying table distances and seen/unseen plate colors, (d) overall performance, and (e) failure cases of each single-frame baseline.}
\label{fig:real_serving}
\end{figure}

\vspace{-.5\baselineskip}
\paragraph{Serving Task} As shown in Fig.~\ref{fig:real_serving}, the robot must navigate while balancing a plate, grasp a cup, and place it upright on the table. This requires stable single-arm transport of the plate during approach, transitioning to bimanual coordination for grasping and local end-effector-centric reasoning for placement. Each policy is trained on 177 demonstrations and evaluated in 20 rollouts across 3 plate colors (1 seen, 2 unseen) and varying table distances. Our~\texttt{MoF-MoE} achieves 70\% success, with four grasping failures and two placement failures. Router weights show \texttt{Left} consistently dominating while \texttt{Right} remains the lowest, indicating that reasoning relative to the stably held plate is optimal throughout the task. Conversely, single-frame baselines exhibit distinct failure modes: (1)~\texttt{Left} (55\%) causes the cup to shift on the plate over extended navigation distances, resulting in grasp failures. (2)~\texttt{Right} (0\%) exhibits highly unstable navigation, failing even to approach the grasp phase. (3)~\texttt{Rel Traj} (60\%) navigates stably but lacks bimanual coordination during the grasping phase.

\section{Conclusion}
\label{sec:conclusion}
We present Mixture of Frames Policy (MoF), a diffusion-policy framework for denoising actions across multiple coordinate-frame parameterizations. Our experiments show that action-frame choice has a substantial and task-dependent effect on diffusion-policy performance, and that fusing frame-specialized denoisers improves over baselines including oracle task-level frame selection.

\section{Limitations}
\label{sec:limitations}
This work has several limitations. First, while theoretically adaptable to larger vision-language-action (VLA) models, our current experiments focus on smaller diffusion policies. Scaling to VLA backbones remains an important future direction. Second, the router is supervised solely via denoising loss, which might not strictly correspond to policy performance. Future work could explore behavior-aware objectives that directly optimize execution quality. Third, MoF relies on a designer-specified set of candidate frames and accurate frame transformations from robot proprioception. Future work could investigate learning or automatically discovering task-relevant frames and study how frame uncertainty affects multi-frame denoising in real-world deployment. Finally, MoF currently relies on frame-dependent Cartesian action representations and does not directly extend to purely joint-space or torque-based policies, whose actions are invariant to coordinate-frame changes.

 
\acknowledgments{
The authors would like to thank Yifan Hou for the help during RB-Y1 controller implementation, and Mengda Xu for the helpful discussion during project brainstorming.
This work utilized Marlowe, Stanford's GPU-based computational instrument~\cite{kapfer2025marlowe}.
This work was supported in part by the  NSF Award \#2143601, \#2037101, and \#2132519, Apple, and Toyota Research Institute. Dian Wang is supported by the Stanford HAI PostDoc fellowship.
The views and conclusions contained herein are those of the authors and should not be interpreted as necessarily representing the official policies, either expressed or implied, of the sponsors.}


\bibliography{reference}  

\clearpage
\input{final_appendix}

\end{document}

%% file: final_appendix.tex
\appendix

\section{Implementation Details}
\label{sec:implementation}

\subsection{Expert Action Parameterizations}
\label{subsec:frames}
We instantiate MoF with four experts that capture complementary action structures in bimanual mobile manipulation: left, right, base-relative (base), and per-arm relative-trajectory (rel traj). The left and right experts use the corresponding end-effector frame, \(F_l\) or \(F_r\), at the most recent observation step. The base-relative expert \(F_b\) represents rotations and translation directions in the mobile-base frame, while re-centering each arm's translation at its observation-step end-effector position. This preserves body- and gravity-aligned structure while reducing absolute translation variance. The relative-trajectory expert \(F_{rt}\), in contrast, represents each arm's future motion in that arm's own current end-effector frame, removing both global translation and global orientation. Thus, the base-relative expert keeps motions aligned with the robot body, while the relative-trajectory expert makes local end-effector-centric motions more compact.

\subsection{MoF-MoE and MoF-Ensemble}
We evaluate two variants of MoF. MoF-MoE, shown in Fig.~\ref{fig:architecture}, uses a router conditioned on the diffuser-conditioning embedding to predict the expert weights \(w_m\). MoF-Ensemble removes the learned router and uses uniform weights, \(w_m=1/|\mathcal{F}|\), effectively averaging the aligned predictions from all frame experts.

\subsection{Per-Expert Auxiliary Loss}
The mixed objective \(\mathcal{L}_{\mathrm{mix}}\) supervises only the fused canonical-frame prediction, so experts with small router weights may receive weak gradients. We therefore add a per-expert auxiliary denoising loss that directly supervises each expert in its own frame. The final training objective is \(\mathcal{L}=\mathcal{L}_{\mathrm{mix}}+\lambda_{\mathrm{aux}}\mathcal{L}_{\mathrm{aux}}\). The full objective is given in Appendix~\ref{app:aux_loss}.

\section{Additional Method Details}
\label{app:method_details}

\subsection{Frame Transformation Operators}
\label{app:transforms}

Here we provide the explicit transformation operators used in Eq.~\eqref{eq:mof_fused_noise}. Let \({}^{m}T_c=({}^{m}R_c,{}^{m}t_c)\) denote the rigid transform from the canonical frame \(F_c\) to an expert frame \(F_m\). For one arm at one action timestep, the canonical-frame action channels are \(x_c=[p_c,c^1_c,c^2_c,g]\), where \(p_c\) is the end-effector position, \(c^1_c,c^2_c\) are the first two rotation columns, and \(g\) is the gripper command.

The action-state transformation \({}^{m}\mathcal{T}_c\) is applied to noisy action states. For each arm and timestep, it maps \(p_c\mapsto{}^{m}R_cp_c+{}^{m}t_c\), \(c^i_c\mapsto{}^{m}R_cc^i_c\), and \(g\mapsto g\). Equivalently, for the vectorized action chunk, \({}^{m}\mathcal{T}_c(x_c)={}^{m}A_cx_c+{}^{m}b_c\), where \({}^{m}A_c\) applies \({}^{m}R_c\) to every position and rotation-column channel and leaves gripper channels unchanged, while \({}^{m}b_c\) inserts the translation \({}^{m}t_c\) only into the position channels.

The denoiser output has the same channel layout, \(\epsilon_c=[\Delta p_c,\Delta c^1_c,\Delta c^2_c,\Delta g]\), but represents a noise vector in the diffusion action space rather than an absolute pose. Therefore, we use the associated linear operator \({}^{m}\mathbf{T}_c\), defined as \({}^{m}\mathbf{T}_c(\epsilon_c)={}^{m}A_c\epsilon_c\). Equivalently, \(\Delta p_c\mapsto{}^{m}R_c\Delta p_c\), \(\Delta c^i_c\mapsto{}^{m}R_c\Delta c^i_c\), and \(\Delta g\mapsto\Delta g\). The inverse operators \({}^{c}\mathcal{T}_m\) and \({}^{c}\mathbf{T}_m\) are defined analogously using the inverse rigid transform \({}^{c}T_m=({}^{m}T_c)^{-1}\).

\subsection{Frame-specific Conditioning}
\label{app:frame_conditioning}

All experts share the same visual features, since RGB observations are not tied to a particular action frame. In contrast, proprioceptive inputs such as end-effector poses and gripper states are transformed into each expert's action parameterization before being provided to that expert. The base-relative and relative-trajectory experts share the same base-frame proprioceptive features.

\subsection{Per-Expert Auxiliary Loss}
\label{app:aux_loss}

The mixed objective \(\mathcal{L}_{\mathrm{mix}}\) supervises only the fused canonical-frame prediction. When the router assigns a small weight to an expert, that expert may receive weak gradients from the mixed loss and drift away from being a meaningful denoiser in its own frame. To stabilize training, we add a per-expert auxiliary loss that directly supervises each expert against the diffusion noise expressed in its own frame.

Given a canonical-frame noisy action \(x^k_c\) and canonical-frame diffusion noise \(\epsilon_c\), expert \(m\) receives \({}^{m}\mathcal{T}_c(x^k_c)\) and predicts noise in frame \(F_m\). The corresponding target is the transformed noise \({}^{m}\mathbf{T}_c(\epsilon_c)\).
The auxiliary loss is
\[
\mathcal{L}_{\mathrm{aux}}(\theta)
=
\mathbb{E}_{k,\epsilon_c,o}
\left[
\sum_{F_m\in\mathcal{F}}
\left\|
{}^{m}\mathbf{T}_c(\epsilon_c)
-
\epsilon^\theta_m({}^{m}\mathcal{T}_c(x_c^k),o;k)
\right\|^2
\right].
\]
The final training objective is \(\mathcal{L}=\mathcal{L}_{\mathrm{mix}}+\lambda_{\mathrm{aux}}\mathcal{L}_{\mathrm{aux}}\). This auxiliary term is used only during training; inference uses the fused noise prediction in Eq.~\eqref{eq:mof_fused_noise}.

\subsection{Difference Between MoE-DP and MoF}
\label{app:moedp_vs_mof}

\begin{figure}[t]
\centering
\includegraphics[width=0.5\linewidth]{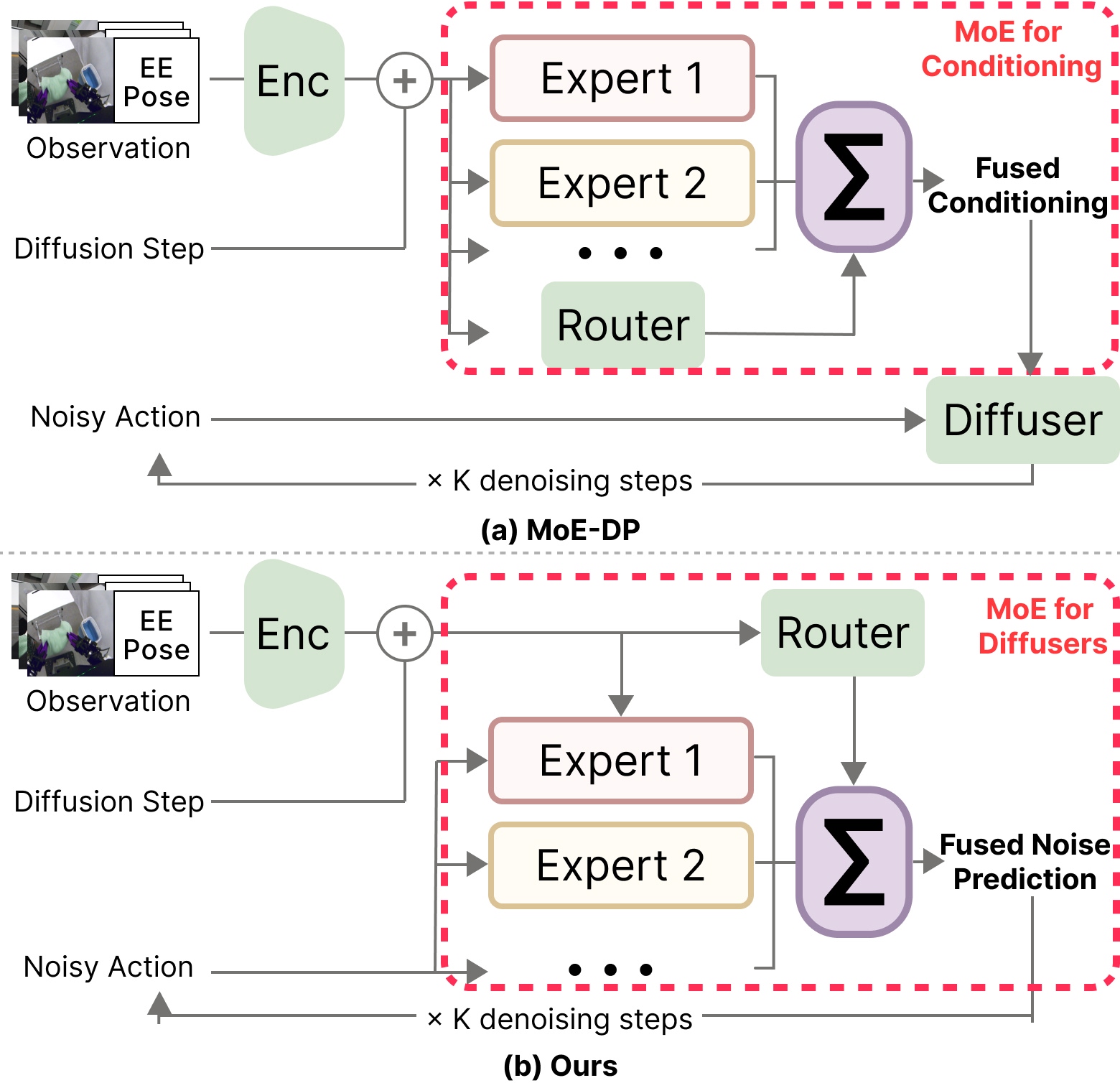}
\caption{\textbf{Difference Between MoE-DP and Our Method.} (a) MoE-DP introduces mixture-of-experts structure in the conditioning pathway, but the diffusion process still denoises actions in a single action representation. (b) MoF instead places the mixture structure directly over frame-specialized denoisers: each expert denoises the same underlying action state expressed in a different coordinate frame. Thus, MoF performs mixture-of-experts reasoning over action frames rather than over conditioning features.}
\label{fig:vs}
\end{figure}

Fig.~\ref{fig:vs} illustrates the difference between a standard mixture-of-experts diffusion-policy baseline and our multi-frame denoising formulation. In MoE-DP, the mixture structure is applied to the conditioning pathway: multiple experts process the observation or conditioning features, their outputs are fused, and a single diffuser denoises actions in one fixed action representation. Therefore, although MoE-DP increases model capacity and allows input-dependent conditioning, the diffusion process itself still operates in a single coordinate frame.

MoF instead places the mixture structure directly over the denoising process. At each diffusion step, the same canonical noisy action is re-expressed in multiple expert frames, each expert predicts noise in its own action representation, and the predicted noises are transformed back to the canonical frame before fusion. Thus, the experts in MoF are not merely alternative conditioning modules; they are frame-specialized denoisers that operate on different coordinate-frame parameterizations of the same underlying action state. This distinction is important for our experiments: the comparison against MoE-DP tests whether the gains come from synchronized multi-frame action denoising rather than from adding a generic MoE module.

\section{Simulation Details}
\label{app:sim_details}

This section provides the full implementation details for our simulation experiments.

\subsection{Benchmarks and Demonstration Data}
\label{app:sim_tasks}

We evaluate on nine bimanual tasks spanning two benchmarks. Five tasks use the BiGym~\cite{chernyadev2025bigym} mobile-manipulation benchmark with the RBY1 bimanual mobile manipulator (\textit{Flip Cup}, \textit{Move 2 Plates}, \textit{Kitchenware}, \textit{Flip Sandwich}, \textit{Dishwasher}), and four tasks use the DexMimicGen~\cite{jiang2025dexmimicgen} tabletop benchmark (\textit{Threading}, \textit{3 Piece Assembly}, \textit{Box Cleanup}, \textit{Drawer Cleanup}).

For both benchmarks, we generate training data by implementing the DexMimicGen~\cite{jiang2025dexmimicgen} data-generation algorithm and extending it to the mobile-manipulation setting when needed. We collect $100$ demonstrations per task for all nine tasks. We follow the task-specific camera configuration defined in each benchmark: two wrist cameras combined with an egocentric head view for the BiGym RBY1 tasks, and two wrist cameras combined with a third-person view for the DexMimicGen tasks.

\subsection{Observation and Action Space}
\label{app:obs_action_space}

All policies share the same observation space and use a common action layout that differs only in the gripper/hand command. The visual observation consists of three RGB camera images (two wrist views and one egocentric/third-person view), each rendered at $84\times84$. All images are preprocessed to $76\times76$ before being passed to the vision encoder; during training we apply a random crop, and at test time a center crop. The low-dimensional observation consists of the two end-effector poses, the proprioceptive gripper/hand state, and the base/head pose; these proprioceptive inputs are re-expressed in each expert's action parameterization before being given to that expert. We use an observation horizon of $2$ steps.

Each arm contributes a $3$D end-effector position and a $6$D rotation representation (the first two columns of the rotation matrix). The five BiGym tasks and the two parallel-jaw DexMimicGen tasks (\textit{Threading}, \textit{3 Piece Assembly}) use a single parallel-jaw gripper per arm, giving a $20$-dimensional action ($9$ per arm plus one gripper command per arm) and a $2$-dimensional gripper proprioception. The two dexterous-hand DexMimicGen tasks (\textit{Box Cleanup}, \textit{Drawer Cleanup}) instead use a $6$-DoF multi-finger hand per arm, giving a $30$-dimensional action ($9$ pose channels plus $6$ finger-joint commands per arm) and a $24$-dimensional hand proprioception ($12$ joints per hand). The MoF frame transformations act only on the per-arm position and rotation channels and leave the gripper/finger channels unchanged, so the same multi-frame denoising procedure applies to both the $20$- and $30$-dimensional actions. Policies predict an action chunk of horizon $16$ and execute the first $8$ predicted actions before re-planning.

\subsection{Training Hyperparameters}
\label{app:sim_hparams}

All policies (MoF and all baselines) share the same diffusion backbone, vision encoder, optimizer, and training schedule; they differ only in the action parameterization and in the method-specific modules described below.

We use the standard Diffusion Policy~\cite{chi2025diffusion} conditional 1D U-Net as the denoiser, conditioned on the observation embedding through FiLM. The U-Net uses channel dimensions $[256, 512, 1024]$, a diffusion-step embedding of dimension $128$, a convolution kernel size of $5$, and $8$ GroupNorm groups. The vision encoder is a ResNet-18, pretrained on ImageNet and fine-tuned end-to-end, with spatial-softmax feature aggregation and GroupNorm; we do not share the encoder across the three cameras. Images are preprocessed to $76\times76$, with a random crop during training and a center crop at test time. We use an observation horizon of $2$ steps, predict an action chunk of horizon $16$, and execute the first $8$ actions before re-planning. The diffusion process uses a DDIM~\cite{ddim} scheduler with $50$ training timesteps and $16$ inference timesteps. 

We optimize all policies with AdamW using a learning rate of $1\times10^{-4}$, weight decay of $1\times10^{-6}$, and $(\beta_1, \beta_2)=(0.95, 0.999)$. The learning rate follows a cosine schedule with $500$ warmup steps. We train with a batch size of $128$ for $500$ epochs.

\subsection{MoF Hyperparameters}
\label{app:sim_mof_hparams}

MoF augments the shared backbone with four frame experts and a learned router (for MoF-MoE). The expert set is $\{\text{base-relative},\,\text{left},\,\text{right},\,\text{rel-traj}\}$, and the base-relative frame is used as the canonical frame $F_c$ for the synchronized denoising trajectory. MoF-MoE uses an MLP router with hidden dimension $512$ that takes the diffuser-conditioning embedding together with the diffusion-step embedding and outputs a softmax distribution over the four experts. MoF-Ensemble replaces the router with fixed uniform weights $w_m=1/|\mathcal{F}|=1/4$. Both variants use the per-expert auxiliary denoising loss of Appendix~\ref{app:aux_loss} with weight $\lambda_{\mathrm{aux}}=1$.

\subsection{Evaluation Protocol}
\label{app:sim_eval}

Each policy is trained for $500$ epochs with three random seeds. For each seed, we evaluate five checkpoints from epochs $460$--$500$ at $10$-epoch intervals, using $50$ rollout episodes per checkpoint. We report the mean success rate over the five checkpoints and three seeds, together with the standard error across the three seeds. Rollout initial configurations use a held-out seed range disjoint from the training demonstrations.

\section{Router Analysis}
\label{app:router_analysis}

We analyze the learned MoF-MoE router across all nine simulation tasks. For each task we load the trained MoF-MoE policy and run it open-loop on $20$ training demonstrations, recording the router's four-expert weight vector at every observation window and every reverse-diffusion step. Throughout this section we focus on the \emph{late denoising phase}, i.e., the per-window weight vector averaged over the last $4$ of the $16$ reverse-diffusion steps. We restrict attention to these clean steps because the router is essentially observation-independent during the high-noise steps (where the action is close to Gaussian noise and provides little signal), and only develops state-dependent structure as the action sharpens.

\subsection{Within-Episode Router Trajectories for All Tasks}
\label{app:router_traj_all}

\begin{figure}[h]
\centering
\includegraphics[width=\linewidth]{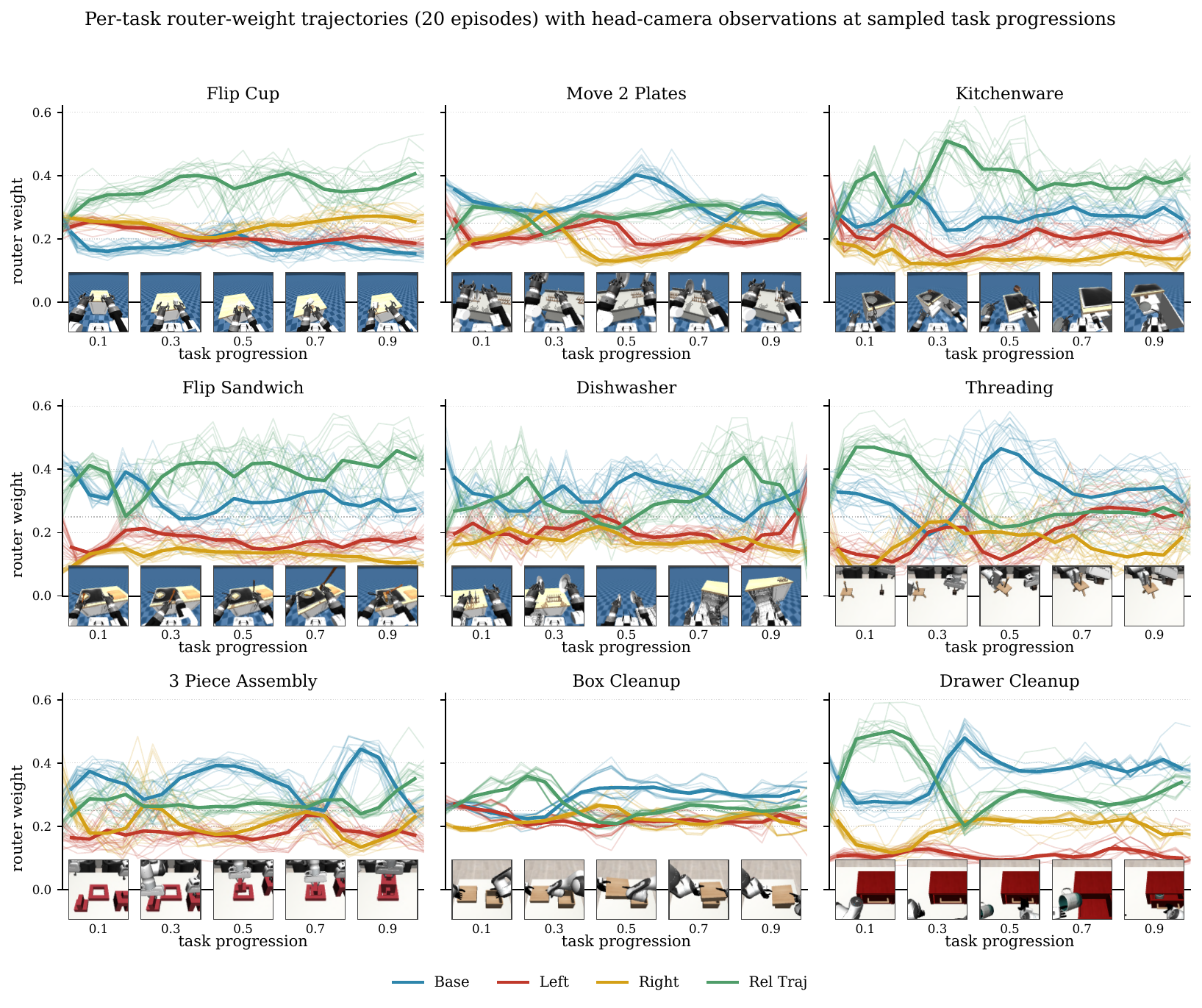}
\caption{\textbf{Router weight trajectories for all nine tasks (20 demonstrations each), with camera observations at sampled task progressions.} For each task, faint lines show the per-episode router weight for each of the four experts (averaged over the last 4 of 16 denoising steps) versus task progression, and bold lines show the across-episode mean; the dotted line at 0.25 marks the uniform-weight reference. Below each panel, camera observations from the demonstration whose router trajectory is closest to the across-episode mean are shown at task progressions 0.1, 0.3, 0.5, 0.7, 0.9. }
\label{fig:router_traj_all}
\end{figure}

Fig.~\ref{fig:router_traj_all} shows the router-weight trajectories for all nine tasks, overlaid with head-camera observations at sampled task progressions. The trajectories vary substantially across tasks in both their static frame preference and their within-episode dynamics. \textit{Threading} and \textit{Drawer Cleanup} show the clearest within-episode handoff: the rel-traj expert dominates the early phase, where each arm independently reaches and grasps its object, and the base expert takes over for the later contact-rich phase that requires coordinated bimanual motion. \textit{Flip Sandwich} and \textit{Kitchenware} keep the rel-traj expert on top throughout with only mild drift, while \textit{3 Piece Assembly}, \textit{Dishwasher}, and \textit{Move 2 Plates} retain a base-dominated soft mixture. \textit{Box Cleanup} is an outlier: all four expert weights cluster tightly near $0.25$ across all phases, indicating a near-uniform router that does not specialize to any frame or phase. \textit{Flip Cup} is notable in the opposite way: the relative ordering of experts is inverted from the other tasks, with the rel-traj expert always on top and the base expert always at the bottom. 

Reading the router trajectories together with the observation frames suggests that the within-episode router dynamics track the visible phase structure of each task. On the tasks with the largest within-episode modulation, the shift from the rel-traj to the base expert coincides with the transition from independent, arm-centric reaching, which is most compact in each arm's own trajectory frame, to coordinated bimanual manipulation, which is most naturally expressed in the shared body frame. This phase-aligned routing is most pronounced precisely on the tasks where MoF-MoE most improves over Oracle Frame (Appendix~\ref{app:router_modulation}), and is essentially absent on \textit{Box Cleanup}, whose near-uniform router cannot exploit such structure. Taken together, this indicates that the learned router contributes most when a task progresses through phases that favor different action frames.

\subsection{Router Modulation Amplitude vs.\ MoF-MoE Advantage over Oracle Frame}
\label{app:router_modulation}

To quantify the within-episode modulation observed in Fig.~\ref{fig:router_traj_all}, we summarize each task by a single scalar that we call the \emph{router modulation amplitude}. We partition each episode's windows into $10$ within-episode time bins, compute the mean late-phase weight per expert per bin, take the standard deviation of each expert's binned-mean weight across the $10$ bins, and average over the four experts. Intuitively, this scalar measures how much each expert's weight typically swings over the course of an episode. 

\begin{figure}[h]
\centering
\includegraphics[width=0.65\linewidth]{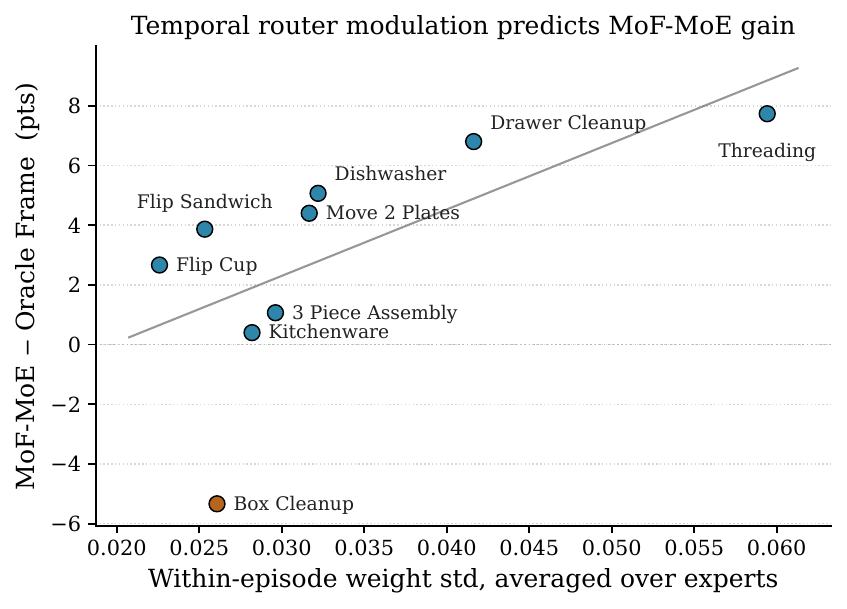}
\caption{\textbf{Router modulation amplitude vs.\ MoF-MoE advantage over Oracle Frame.} Each point is one task. The $x$-axis is the within-episode standard deviation of each expert's binned-mean weight, averaged over the four experts. The $y$-axis is the difference between MoF-MoE's and Oracle Frame's last-$5$ success rate. The orange marker (\textit{Box Cleanup}) is the only task where MoF-MoE underperforms Oracle Frame and is also the task with relatively low router modulation amplitude. Spearman $\rho=+0.75$, $p=0.020$, $n=9$.}
\label{fig:time_div_vs_advantage}
\end{figure}

Fig.~\ref{fig:time_div_vs_advantage} plots the router modulation amplitude against MoF-MoE's per-task advantage over Oracle Frame. The two quantities are positively correlated across tasks (Spearman $\rho=+0.75$, $p=0.020$, $n=9$). The two tasks with the largest router modulation, \textit{Threading} and \textit{Drawer Cleanup}, are also the tasks where MoF-MoE most improves over the best single frame ($+7.7$ and $+6.8$ points). At the other end of the spectrum, \textit{Box Cleanup} has a relatively low router modulation amplitude---its router is close to uniform throughout each episode---and is the only task where MoF-MoE underperforms Oracle Frame. \textit{Move 2 Plates}, \textit{Dishwasher}, \textit{Flip Sandwich}, and \textit{Flip Cup} show smaller gains over Oracle Frame than Threading and Drawer Cleanup. This relationship suggests that MoF-MoE's advantage over the best single frame stems primarily from its ability to assign different frames to different task phases, rather than from a static preference for any one frame.

\section{Ablation Study Details}
\label{app:ablation}

We provide the full ablation results for the four representative tasks used in Sec.~\ref{sec:sim_exp}: \textit{FlipCup}, \textit{Move2Plates}, \textit{Threading}, and \textit{BoxCleanup}. These tasks cover both BiGym and DexMimicGen environments and include settings where different action parameterizations are favored. Parenthetical values denote differences from the corresponding baseline: MoF-MoE for MoE variants and MoF-Ensemble for the ensemble variant.

\begin{table}[h]
\centering
\caption{Full ablation study for MoF. Last-5 success rate (\%, mean $\pm$ std.\ err.\ over 3 seeds). Parenthetical values denote differences from the corresponding baseline.}
\label{tab:mof_ablation_full}
\scriptsize
\setlength{\tabcolsep}{2.5pt}
\begin{tabular}{l c c c c c}
\toprule
Method & Avg & Flip Cup & Move 2 Plates & Threading & Box Cleanup \\
\midrule
MoF MoE (Ours) & 66.5 & $56.5\pm1.4$ & $51.6\pm4.4$ & $70.8\pm7.5$ & $87.1\pm3.2$ \\
\midrule
\quad Rel Traj as Canonical & 64.0 ($-2.5$) & $56.8\pm3.6$ ($+0.3$) & $45.9\pm3.3$ ($-5.7$) & $67.6\pm2.6$ ($-3.2$) & $85.7\pm5.3$ ($-1.4$) \\
\quad Left as Canonical & 65.8 ($-0.7$) & $56.8\pm2.0$ ($+0.3$) & $51.3\pm3.8$ ($-0.3$) & $67.6\pm1.2$ ($-3.2$) & $87.6\pm6.2$ ($+0.5$) \\
\quad Right as Canonical & 66.4 ($-0.1$) & $55.3\pm2.6$ ($-1.2$) & $54.7\pm6.1$ ($+3.1$) & $71.3\pm2.2$ ($+0.5$) & $84.3\pm1.4$ ($-2.8$) \\
\midrule
\quad w/o Best Frame & 60.3 ($-6.2$) & $54.9\pm2.9$ ($-1.6$) & $45.1\pm5.3$ ($-6.5$) & $63.2\pm5.0$ ($-7.6$) & $77.9\pm8.6$ ($-9.2$) \\
\quad w/o Aux & 58.5 ($-8.0$) & $45.5\pm2.9$ ($-11.0$) & $30.8\pm0.9$ ($-20.8$) & $70.3\pm6.1$ ($-0.5$) & $87.5\pm5.3$ ($+0.4$) \\
\quad w/ Ortho Proj & 59.5 ($-7.0$) & $49.2\pm2.1$ ($-7.3$) & $38.4\pm3.8$ ($-13.2$) & $64.4\pm8.9$ ($-6.4$) & $85.9\pm1.4$ ($-1.2$) \\
\midrule
MoF Ensemble (Ours) & 64.2 & $59.2\pm0.4$ & $51.5\pm1.5$ & $68.7\pm3.3$ & $77.5\pm5.2$ \\
\midrule
\quad w/ Ortho Proj & 0.0 ($-64.2$) & $0.0\pm0.0$ ($-59.2$) & $0.0\pm0.0$ ($-51.5$) & $0.0\pm0.0$ ($-68.7$) & $0.0\pm0.0$ ($-77.5$) \\
\bottomrule
\end{tabular}
\end{table}

Changing the canonical frame has only a modest effect. Using relative-trajectory, left, or right as the canonical frame changes the average success rate by at most 2.5 percentage points, suggesting that MoF is not sensitive to the arbitrary choice of scheduler frame. Removing the best-performing single frame from the expert set reduces the average success rate by 6.2 percentage points, showing that the expert set is not redundant and that task-relevant frames contribute complementary denoising structure.

The orthogonalization ablation validates the transformation-compatible action representation. When noisy rotation channels are projected back to \(\SO(3)\) before applying frame transformations, MoF-MoE drops by 7.0 percentage points on average, and MoF-Ensemble collapses completely. This is because the orthogonalization completely destroys all non-canonical-frame experts. Although MoF-MoE is able to escape by assigning all weights to the canonical expert, there is no way for MoF-Ensemble to recover from this failure. This supports our design choice of using column-vector rotation channels, which can be transformed directly as ordinary 3D vectors without changing the noisy diffusion state.

Finally, removing the per-expert auxiliary denoising loss reduces average performance by 8.0 percentage points. This suggests that directly supervising each expert in its own frame is important for keeping all frame experts meaningful during training, especially when the fused loss may provide weak gradients to experts with small router weights.

\section{Expert Frame Selection and Scalability}
\label{app:frame_scalability}

We investigate how the number and choice of expert frames affect MoF performance and computational cost. We evaluate MoF-MoE on four BiGym tasks (\textit{Flip Cup}, \textit{Move 2 Plates}, \textit{Kitchenware}, and \textit{Flip Sandwich}), comparing the best single-frame policy, a three-expert variant excluding the oracle frame, and our default four-expert configuration. We further augment the default expert set with five additional frames: mid-torso, egoview, shoulder, elbow, and a virtual end-effector midpoint. We also measure training and inference latency and GPU memory usage as the number of experts increases.

\begin{table}[t]
\centering
\caption{Expert frame scalability: last-5 success rate (\%, mean $\pm$ std.\ err. over 3 seeds). Bold indicates best.}
\label{tab:frame_scaling}
\resizebox{\columnwidth}{!}{%
\begin{tabular}{lcccc}
\toprule
Frame set & Flip Cup & Move 2 Plates & Kitchenware & Flip Sandwich \\
\midrule
1F DP (oracle frame) & $53.9\pm2.9$ & $47.2\pm1.7$ & $31.2\pm2.6$ & $36.1\pm3.3$ \\
3F MoE (w/o oracle frame) & $54.9\pm2.9$ & $45.1\pm5.3$ & $34.5\pm1.3$ & $37.7\pm2.9$ \\
\textbf{4F MoE (Ours)} & $56.5\pm1.4$ & $51.6\pm4.4$ & $31.6\pm2.4$ & $\mathbf{40.0\pm3.1}$ \\
\midrule
4F + mid-torso-relative & $55.7\pm5.0$ & $47.5\pm2.3$ & $32.9\pm2.2$ & $36.8\pm0.6$ \\
4F + egoview-relative & $57.7\pm1.9$ & $53.2\pm1.5$ & $28.4\pm2.8$ & $39.5\pm2.4$ \\
4F + shoulder-relative & $\mathbf{58.0\pm2.3}$ & $46.4\pm2.7$ & $28.9\pm4.1$ & $36.8\pm1.4$ \\
4F + elbow-relative & $56.9\pm1.1$ & $44.1\pm2.3$ & $29.3\pm2.2$ & $33.9\pm2.1$ \\
4F + EE-midpoint-relative (virtual) & $55.6\pm0.9$ & $\mathbf{53.7\pm3.0}$ & $\mathbf{34.8\pm1.3}$ & $32.1\pm1.9$ \\
\bottomrule
\end{tabular}%
}
\end{table}

Table~\ref{tab:frame_scaling} shows that removing the oracle frame degrades performance on average, while adding a fifth frame improves some tasks but provides no consistent gains across tasks. Computational costs scale approximately linearly with the number of experts, with training and inference latency increasing by 22.1 and 58.5 ms per expert, and GPU memory by 1.94 and 0.33 GiB, respectively, on an RTX PRO 6000 Blackwell. Thus, adding more frames incurs additional computational cost without guaranteed performance gains. Our default four-frame configuration provides a natural, compact set of complementary reference frames for bimanual mobile manipulation without requiring task-specific or embodiment-specific frame design.

\section{Real-World Experiment Details and Analysis}
\label{app:real_details}

\subsection{Task Definitions}
\label{app:real-world-task-definitions}

We evaluate two real-world tasks: pouring and serving. In the \textbf{pouring task}, the robot picks up two cups, transfers a red cube from the right cup to the left cup, and places both cups upright on the table. Success requires the cube to end in the left cup and the robot to release both cups while they remain upright. The initial configurations vary the ordered right/left cup-color pair and the table placement distance from the robot over four distance levels. The seen color pairs are (Yellow, Gray), (Yellow, Blue), and (Gray, Blue), while the unseen color pairs are (Blue, Purple) and (Purple, Gray). In the \textbf{serving task}, the robot navigates toward the table, picks up a cup, and places it upright on the table. Success requires the cup to be placed upright without falling or being released incorrectly. We vary the plate color and the distance between the robot's initial base pose and the table. Purple plates are seen during training and are evaluated at six base-to-table distances, while Blue and Milk plates are unseen and are evaluated at seven distances, including one additional closest-distance setting.

\subsection{Training Details}
\label{app:real-world-training-details}

We integrate our MoF policy within the HoMMI~\cite{xu2026hommi} framework by implementing our column-vector SE(3) action representation and frame transformations, while keeping the HoMMI runtime interface unchanged. The MoF-MoE policy uses left, right, and relative-trajectory experts, with the left frame as the canonical frame. Since the HoMMI data collection interface does not record the base frame, we disable the base frame and exclude the base-relative expert. For the relative-trajectory expert, low-dimensional observations are therefore expressed using the left-frame observation variant. The head look-at action is represented as a 3D point rather than a pose, so it contains no orientation channel and is transformed using point-frame transformations. For the relative-trajectory expert, this look-at point is also expressed in the left frame. We train real-world MoF-MoE with the same objective as in simulation, including the mixed canonical denoising loss and the per-expert auxiliary denoising loss.

All evaluated policies use a DiT-based diffusion-policy backbone. The policies use an observation horizon of 2, an action horizon of 32, represent rotations with 6D rotation channels, and predict a 23-dimensional action vector. The action vector consists of two end-effector target poses, each represented by a 3D position and a 6D rotation, two gripper commands, and a 3D head look-at point. The DDIM scheduler uses 100 training timesteps, a squared-cosine beta schedule, and epsilon prediction. During inference, we use 16 denoising steps. Visual preprocessing follows the training pipeline. Head images are resized to $224 \times 224$, wrist images are cropped and resized to $96 \times 96$, and the pointmap branch samples 512 points from the head pointmap. Training augmentation applies color jitter to the head stream. For wrist streams, we apply center crop, random crop, color jitter, and resize. We train with a policy learning rate of $7.5\times10^{-5}$, a visual-encoder learning rate of $7.5\times10^{-6}$, weight decay of $10^{-6}$, a cosine learning-rate schedule, 50 warmup steps, EMA, batch size 16, and gradient accumulation of 2. We evaluate the epoch-500 checkpoints for serving and the epoch-580 checkpoints for pouring.

\subsection{Deployment Details}
\label{app:real-world-deployment-details}

We adopt the HoMMI~\cite{xu2026hommi} hardware setup and whole-body controller for all real-world rollouts. The policy server runs at 5 Hz with an action horizon of 32. The controller executes policy trajectories at 5 Hz, while state estimation and IK run at 100 Hz. A new policy segment is requested every 3.6 s for serving and every 3.2 s for pouring. The controller linearly interpolates the received policy targets and applies a 1 Hz low-pass filter to the commands. For safety and compliance during real-world execution, we enable both the compliant controller and the end-effector wrench safety guard. The admittance controller uses zero desired wrench, translational stiffness $(300,300,300)$, rotational stiffness $(1,1,1)$, translational damping $(7,7,7)$, and rotational damping $(0.2,0.2,0.2)$. We use translation-force mode with direct force-control gains set to zero. The wrench safety guard uses calibrated end-effector wrench readings and latches the output when the measured force exceeds $120~\mathrm{N}$ or the measured torque exceeds $12~\mathrm{Nm}$.

The same whole-body IK parameters are used for both serving and pouring. The interpolated Cartesian targets are converted into generalized joint velocities using the HoMMI differential whole-body IK solver. We use the \texttt{daqp} optimizer with a damping coefficient of $10^{-6}$ for numerical stability. The IK objective prioritizes accurate bimanual end-effector tracking, with equal position and orientation tracking weights of $10000$. We regularize the robot toward a nominal torso posture with weight $200$, while disabling nominal arm-posture regularization. To smooth the solution in velocity space, we penalize deviations from the current configuration with weight $10$ for the joints. The mobile base is regularized more strongly, with both translational and rotational base components weighted by $5000$. Unlike the original HoMMI controller, we disable the center-of-mass-over-base objective, while keeping the relative torso--base displacement bounds at $0.15~\mathrm{m}$ in both the $x$ and $y$ directions.

\subsection{Failure Mode Analysis}
\label{app:real-world-failure-mode-analysis}

In the pouring task, \textbf{MoF-MoE} achieves 85\% success, with only three failures: two right-cup pickup failures and one left-cup pickup failure, all on seen cup-color combinations. Importantly, in all 17 rollouts where MoF-MoE successfully picks up both cups, it completes both cube transfer and upright placement, suggesting that multi-frame fusion primarily improves robustness in the initial grasping phase while preserving reliable bimanual coordination afterward. The single-frame baselines fail in distinct, frame-dependent ways. The \textbf{Left} policy achieves only 15\% success, with 12 right-cup pickup failures, two left-cup pickup failures, two pouring failures, and one rollout that repeatedly continues the pouring motion even after successful transfer. It fails on all unseen color combinations, and its errors are dominated by the right arm. Conversely, the \textbf{Right} policy achieves 70\% success, but all of its failures are left-side failures: four left-cup pickup failures and two left-cup placement failures, all on seen color combinations. These left- and right-frame results suggest that when the policy is trained in a single end-effector frame, predictions for the opposite arm become insufficiently precise. The \textbf{Rel Traj} policy achieves only 5\% success and is highly unstable during precise pickup, with 13 left-cup pickup failures, three failures to pick up both cups, two transfer failures due to insufficient bimanual coordination, and one right-cup placement failure. This indicates that relative-trajectory actions alone are not stable enough for the fine-grained grasping required by pouring.

In the serving task, \textbf{MoF-MoE} achieves 70\% success, with four cup-pickup failures and two placement failures. One failure occurs on seen plate-color configurations and five occur on unseen plate-color configurations. The failures are concentrated at the closest and intermediate base-to-table distances, while MoF-MoE succeeds on all configurations that require longer-distance navigation, indicating stable navigation over extended approach motions. The \textbf{Left} policy achieves 55\% success and fails nine times at cup pickup. More than half of these failures occur at farther-than-average base-to-table distances, suggesting that long-distance navigation causes the cup to shift on the plate and leads to subsequent pickup failure. The \textbf{Right} policy fails in all 20 rollouts and exhibits highly unstable navigation throughout, suggesting that using the right frame as the sole reference induces a wider and harder-to-learn action distribution for this task. The \textbf{Rel Traj} policy achieves 60\% success, with all eight failures occurring during the cup-pickup phase due to insufficient bimanual coordination. In contrast to the Left policy, more than half of the relative-trajectory failures occur at closer-than-average base-to-table distances, indicating that relative-trajectory actions support stable navigation but do not provide the coordinated bimanual reference needed for grasping the cup from the plate.